\documentclass[twocolumn,amsmath,nofootinbib]{revtex4} 
\usepackage{url}
\usepackage{amssymb}
\usepackage{amsfonts}
\usepackage{amsbsy}
\usepackage{hyperref}
\usepackage{amssymb}
\usepackage{mathrsfs}
\usepackage{amsmath}
\usepackage{mathtools}
\usepackage{algorithm}
\usepackage{algorithmic}
\usepackage{tabularx}
\usepackage{enumitem}
\usepackage{mathtools}
\usepackage{empheq}

\def\ie{{\frenchspacing\it i.e.}}
\def\eg{{\frenchspacing\it e.g.}}
\def\etc{{\frenchspacing\it etc.}}
\def\rms{{\frenchspacing r.m.s.}}
\def\expec#1{\langle#1\rangle}

\def\A{\textbf{A}}

\def\J{\textbf{J}}

\def\Ell{\mathcal{L}}
\def\M{\textbf{M}}

\def\g{\textbf{g}}

\def\rvec{\textbf{r}}

\def\a{\textbf{a}}

\DeclareMathAlphabet\mathbfcal{OMS}{cmsy}{b}{n}

\def\ie{{\frenchspacing\it i.e.}}
\def\eg{{\frenchspacing\it e.g.}}
\def\etc{{\frenchspacing\it etc.}}
\def\rms{{\frenchspacing r.m.s.}}
\def\expec#1{\langle#1\rangle}

\def\A{\textbf{A}}

\def\I{\textbf{I}}
\def\J{\textbf{J}}

\def\Ell{\mathcal{L}}
\def\M{\textbf{M}}

\def\a{\textbf{a}}

\def\g{\textbf{g}}

\def\rvec{\textbf{r}}

\def\w{\textbf{w}}
\def\x{\textbf{x}}

\def\z{\textbf{z}}

\def\zero{\textbf{0}}

\usepackage{amsmath}

 \def\logplus{\log_+}

\def\spose#1{\hbox to 0pt{#1\hss}}
\def\simlt{\mathrel{\spose{\lower 3pt\hbox{$\mathchar"218$}}
   \raise 2.0pt\hbox{$\mathchar"13C$}}}
\def\simgt{\mathrel{\spose{\lower 3pt\hbox{$\mathchar"218$}}
     \raise 2.0pt\hbox{$\mathchar"13E$}}}
 \def\simpropto{\mathrel{\spose{\lower 3pt\hbox{$\mathchar"218$}}
     \raise 2.0pt\hbox{$\propto$}}}

\def\beq#1{\begin{equation}\label{#1}}
\def\eeq{\end{equation}}
\def\beqa#1{\begin{eqnarray}\label{#1}}
\def\eeqa{\end{eqnarray}}
\def\eq#1{equation~(\ref{#1})}

\def\fig#1{Figure~\ref{#1}}
\def\Fig#1{Figure~\ref{#1}}

\def\Sec#1{Section~\ref{#1}}

\parskip=\medskipamount
\begin{document}
\title{Symbolic Pregression: Discovering Physical Laws from Distorted Video}
\author{Silviu-Marian Udrescu, Max Tegmark}
\address{Dept.~of Physics \& Center for Brains, Minds \& Machines, Massachusetts Institute of Technology, Cambridge, MA 02139; sudrescu@mit.edu}
\begin{abstract}
We present a method for unsupervised learning of equations of motion for objects in raw and optionally distorted unlabeled synthetic video.
We first train an autoencoder that maps each video frame into a low-dimensional latent space where the laws of motion are as simple as possible, by minimizing a combination of non-linearity, acceleration and prediction error. 
Differential equations describing the motion are then discovered using Pareto-optimal symbolic regression.
We find that our pre-regression (``pregression'') step is able to rediscover Cartesian coordinates of unlabeled moving objects even when the video is distorted 
by a generalized lens.
Using intuition from multidimensional knot-theory, we find that the 
pregression step is facilitated by first adding extra latent space dimensions to avoid topological problems during training and then removing these extra dimensions via principal component analysis.
An inertial frame is auto-discovered by minimizing the combined equation complexity for multiple experiments. 
\end{abstract}
\date{\today}
\vspace{10mm}	

\maketitle

\section{Introduction}
\label{IntroSec}

A central goal of physics and science more broadly is to discover mathematical patterns in data. For example, after four years of analyzing data tables on planetary orbits, Johannes Kepler started a scientific revolution in 1605 by discovering that 
Mars' orbit was an ellipse \cite{koyre2013astronomical}.
There has been great recent progress in automating such tasks with {\it symbolic regression}: 
discovery of a symbolic expression that accurately matches a given data set \cite{crutchfield1987equation, dzeroski1995discovering, bradley2001reasoning, langley2003robust, schmidt2009distilling, mcree2010symbolic, searson2010gptips, dubvcakova2011eureqa, stijven2011separating, schmidt2011automated, hillar2012comment, daniels2015automated, langley2015heuristic, arnaldo2015building, brunton2016discovering, guzdial2017game, quade2018sparse, koch2018mutual, kong2018new, liang2019phillips, udrescu2020ai, udrescu2020ai2}. Open-source software now exists that can discover quite complex physics equations by combining neural networks with techniques inspired by physics and information theory \cite{udrescu2020ai,udrescu2020ai2}.

However, there is an important underlying problem that symbolic regression does not solve: how to decide which parameters of the observed data we should try to describe with equations.
Eugene Wigner famously stated that  ``the world is very complicated and ... the complications are called initial conditions, the domains of regularity, laws of nature" \cite{wigner1949invariance}, so how can the discovery of these regularities be automated?
In \fig{QuarticFig}, how can an unsupervised algorithm learn that to predict the next video frame, it should focus on the $x$- and $y$-coordinates of the rocket, not on its color or on the objects in the background? 
More generally, given an evolving data vector with $N$ degrees of freedom, how can we auto-discover which 
$n<N$ degrees of freedom are most useful for prediction? Renormalization addresses this question in a particular context, 
but we are interested in generalizing this.
In Kepler's case, for example, the raw data corresponds to two-dimensional sky images:
how can a computer algorithm presented with a series of images with say $N=10^6$ pixels automatically learn that the most useful degrees of freedom are the $n=2$ position coordinates of the small reddish dot corresponding to Mars?

The goal of our paper is to tackle this pre-regression problem, which we will refer to this  as {\it ``pregression"} for brevity. 
Automated pregression enables laws of motion to be discovered starting with raw observational data such as videos.
This can be viewed as a step toward unsupervised learning of physics, whereby 
an algorithm learns from raw observational data without any human supervision or prior knowledge \cite{lecun2015deep, wu2019toward, iten2018discovering}.

\onecolumngrid

\begin{figure}[h] 
\centerline{\includegraphics[width=0.86\linewidth]{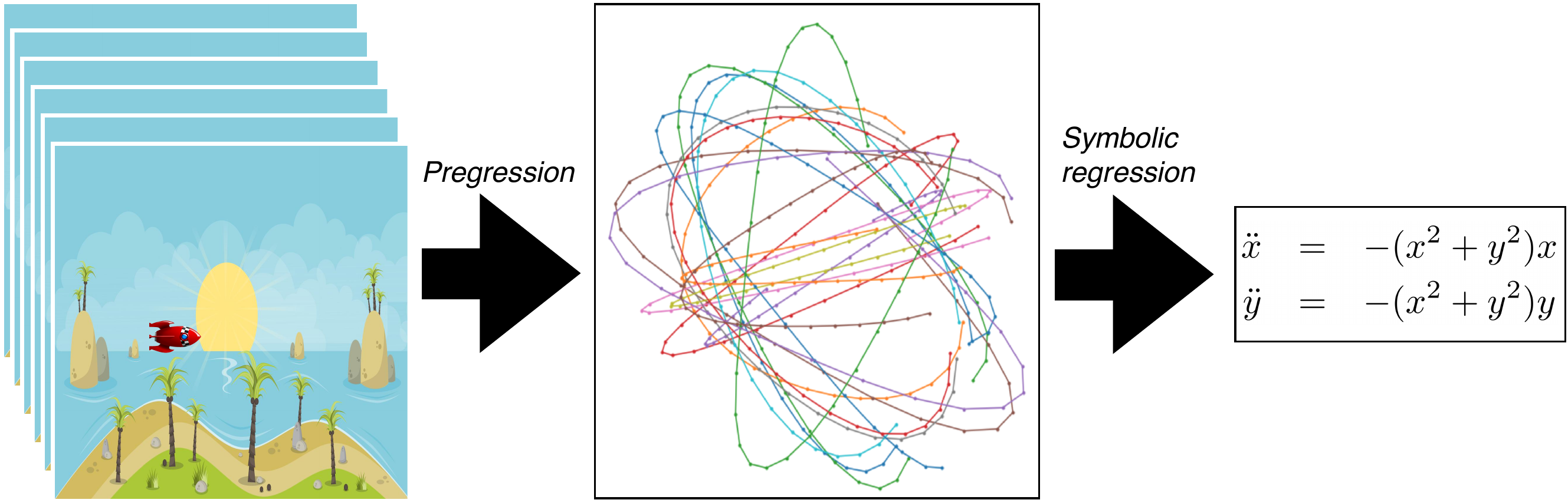}}
\end{figure} 
\vskip-1.4mm
{\footnotesize FIG. 1: Our pregression algorithm seeks to autoencode a sequence of video frames (left) corresponding to a specific type of motion into a low-dimensional latent space (middle) where the laws of motion (right) are as simple as possible, in this example those of a quartic oscillator. In the middle figure each point corresponds to the x and y of the rocket in a given frame, while points having the same color and connected by a line belong to the same trajectory.
\vskip5mm

\twocolumngrid

There has been impressive recent progress on using neural networks 
for video prediction \cite{ranzato2014video, michalski2014modeling, srivastava2015unsupervised, oh2015action, finn2016unsupervised, lotter2016deep, cricri2016video, kalchbrenner2017video, liang2017dual, denton2017unsupervised, villegas2017decomposing, vukotic2017one, babaeizadeh2017stochastic, oliu2018folded, liu2019deep} and more general physics problems  \cite{carrasquilla2017machine, van2017learning, van2018learning, iten2018discovering, torlai2016learning, ohtsuki2017deep, dunjko2018machine}.
However, these machine-learned models tend to be inscrutable black boxes that provide their human users with limited understanding.
In contrast, the machine learning approach in this paper aspires to {\it intelligible intelligence}, \ie, learning a model of the system that is simple enough for a human user to understand.
Such intelligibility (pursued in, \eg, \cite{yildirim2018neurocomputational, zheng2018unsupervised, chang2016compositional, wu2019toward, iten2018discovering, zhang2019general})
is a central goal of physics research, and 
has two advantages:
\begin{enumerate}
\itemsep0mm
\item Understanding how a model works enables us to trust it more, which is particularly valuable when AI systems make decisions affecting peoples lives \cite{russell2015research, amodei2016concrete, boden2017principles, battaglia2018relational}.
\item Simple intelligible models such as the laws of physics tend to yield more accurate and generalizable predictions than black-box over-parametrized fits, especially over long timescales. This is why spacecraft navigation systems use Newton's law of gravitation rather than a neural-network-based approximation thereof.
\end{enumerate}

The video prediction papers most closely related to the present work take one of two approaches. Some improve accuracy and intelligibility by hardcoding coordinate-finding or physics elements by hand to help learn  {\eg}
rigid-body motion \cite{bhat2002computing},
physical object properties or
partial differential equations \cite{tompson2017accelerating,lu2019extracting}.
The alternative {\it tabula rasa} 
approach assumes no physics whatsoever and attempts
to learn physical object properties \cite{wu2016physics}, 
object positions \cite{santoro2017simple,liu2018intriguing},
object relations \cite{hamrick2018relational}
and time evolution \cite{watters2017visual,oord2018representation,greydanus2019hamiltonian}
by learning a low-dimensional representation or latent space which is unfortunately too complex or inscrutable to allow discovery of exact equations of motion.
The present paper builds on this {\it tabula rasa} approach; our key contribution is to automatically simplify the latent space, using ideas inspired by general relativity and knot theory, to make the dynamics simple enough for symbolic regression to discover equations of motions.

The rest of this paper is organized as follows. 
In \Sec{MethodsSec}, we present our algorithm.
In  \Sec{ResultsSec}, we test it on simulated videos (such as the flying rocket example in \fig{QuarticFig}) 
for 
motion 
in a force-free environment, 
a gravitational field, 
a magnetic field, 
a harmonic potential and
a quartic potential. We also test the effects of adding noise and geometric image distortion.
We summarize our conclusions and discuss future challenges in \Sec{ConclusionsSec}.

\section{Method}
\label{MethodsSec}

The goal of our method is to start with raw video sequences of an object moving in front of some static background, and to, in a fully unsupervised manner (with no input besides the raw video) discover the differential equation governing the object's motion. Our algorithm consists of two parts:
\begin{enumerate}
\itemsep0mm
\item a neural-network-based pregression step that learns to map images into a low-dimensional latent space representing the physically relevant parameters (degrees of freedom), and
\item a symbolic regression step that discovers the law of motion, \ie, the differential equation governing the time-evolution of these parameters.
\end{enumerate}

\subsection{Learning the latent space}

\begin{figure}[phbt]
\centerline{\includegraphics[width=\linewidth]{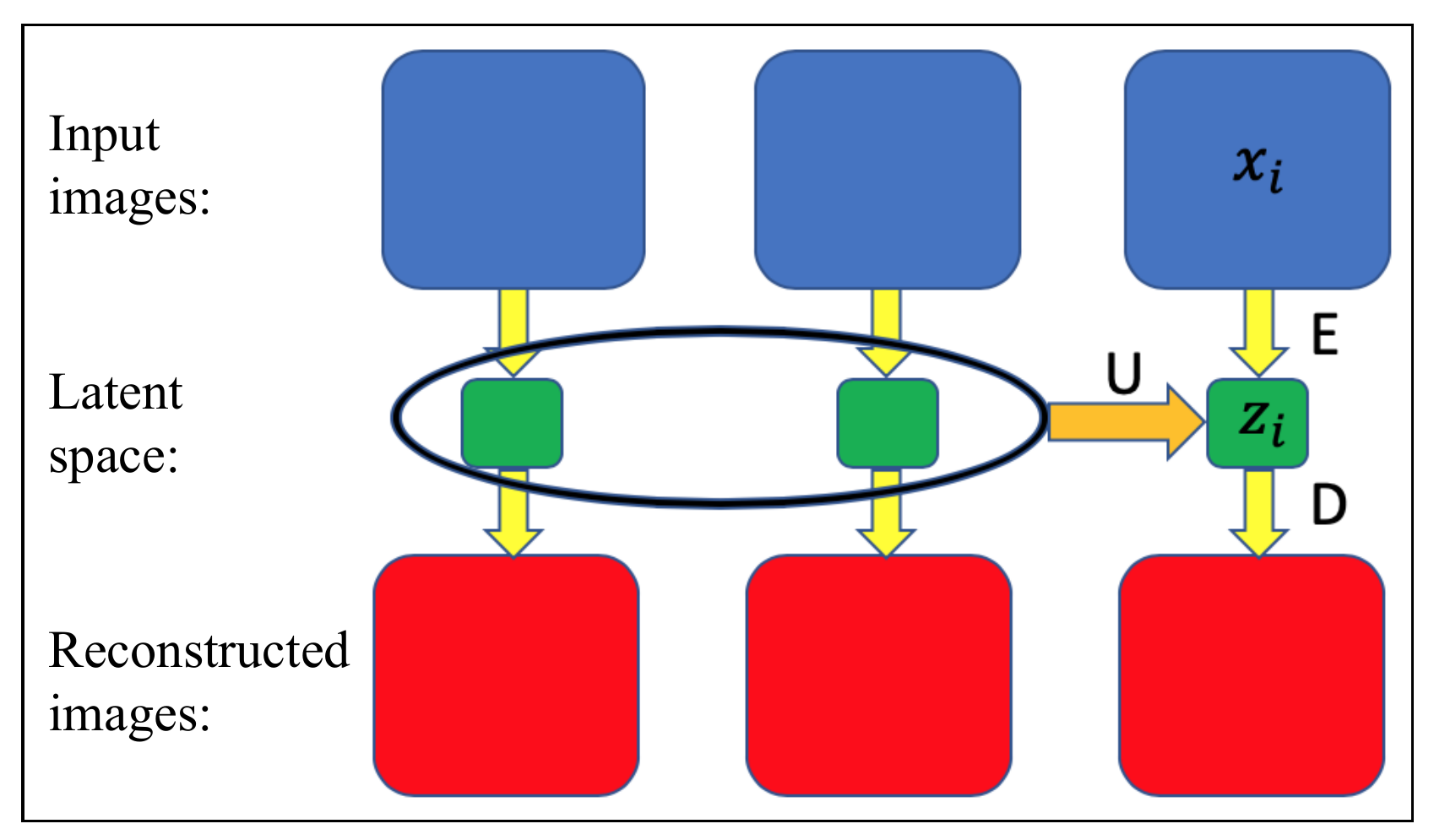}}
\caption{
\protect\setcounter{figure}{0}
\protect\label{QuarticFig}
\protect\setcounter{figure}{2}
Architecture of our neural network:
an encoder $E$ that maps images $\x_i$ into latent space vectors $\z_i$,
a decoder $D$ that maps  latent space vectors $\z_i$ back into images $\x_i$, 
and an evolution operator $U$ that predicts the next latent space vector from the two previous ones.
\label{architectureFig}
}
\end{figure}

Abstractly, we can consider each video frame as a single point in an $N$-dimensional space, where $N$ is the number of color channels (3 in our case) times the number of pixels in each image. If the motion involves only $n\ll N$ degrees of freedom 
(for example, $n=2$ for a rigid object moving without rotating in two dimensions), then 
all observed points in the $N$-dimensional space lie on some $n$-dimensional submanifold that we wish to discover, parametrized by an $n$-dimensional parameter vector that we can consider as a point in an $n$-dimensional latent space.
Our neural network architecture for learning the latent space is shown in \Fig{architectureFig}, and consists of three separate feedforward neural networks:
\begin{enumerate}
\itemsep0mm
\item An encoder $E$ that maps images $\x_i\in\mathbb{R}^{N}$ into latent space vectors $\z_i\in\mathbb{R}^{n}$,
\item a decoder $D$ that maps  latent space vectors $\z_i$ into images $\x_i$, and
\item an evolution operator $U$ that predicts the next latent space vector $\z_i$ from the two previous ones (two are needed to infer velocities).\footnote{The two last images are needed because the laws of physics are second order differential equations that can be transformed into second order difference equations; our method trivially generalizes to using the last $T$ inputs for any choice $T=1, 2, 3, ...$}
\end{enumerate}
The encoder-decoder pair forms an autoencoder 
\cite{bourlard1988auto, lecun1989backpropagation, hinton1994autoencoders, hinton2006reducing, bengio2012unsupervised, kingma2013auto, bengio2013deep, higgins2017beta, achille2018information} that tries to 
discover the $n$ most dynamically relevant parameters from each movie frame, from which it can be reconstructed as accurately as possible.
 
\subsection{Quantifying simplicity}
\label{LossSec}

It is tempting to view the results of our pregression algorithm as rather trivial, merely learning to extract $x-$ and $y-$ coordinates of objects.
This would be incorrect, however, since we will see that the pregression discovers simple physical laws even from video images that
are severely warped, as illustrated in \fig{warpingFig}, where the learned latent space is a complicated non-linear function of the Cartesian coordinates. The basic reason for this is that \fig{architectureFig} makes no mention of any preferred latent-space coordinate system.
This reparametrization invariance (a core feature of general relativity) is a double-edged sword, however:
a core challenge that we must overcome is that even if the system {\it can} be described by a simple time-evolution $U$, the basic architecture in \fig{architectureFig} may discover something much more complicated. 
To see this, suppose that there is an autoencoder $(E,D)$ and evolution operator $U$ providing perfect image reconstruction and prediction, \ie, satisfying
\begin{equation}
\begin{split}
&D(E(\x_i))=\x_i,\\
&U(\z_{i-2},\z_{i-1})=\z_i
\end{split}
\end{equation}
and that $U$ is a fairly simple function.
If we now deform the latent space by replacing $\z$ by $\z'\equiv f(\z)$ for some invertible but horribly complicated function $f$, then it is easy to see that the new mappings defined by 
\begin{equation}
\begin{split}
E'(\x)&\equiv f(E(\x)),\\
D'(\z')&\equiv D(f^{-1}(\z')), \\
U'(\z')&\equiv f(U(f^{-1}(\z')))
\end{split}
\end{equation}
will still provide perfect autoencoding and evolution
\begin{equation}
\begin{split}
&D'(E'(\x_i))=\x_i,\\
&U'(\z_{i-2},\z_{i-1}))=\z_i
\end{split}
\end{equation}
even though the new evolution operator $U'$ is now very complicated.

\begin{figure}[phbt]
\centerline{\includegraphics[width=\linewidth]{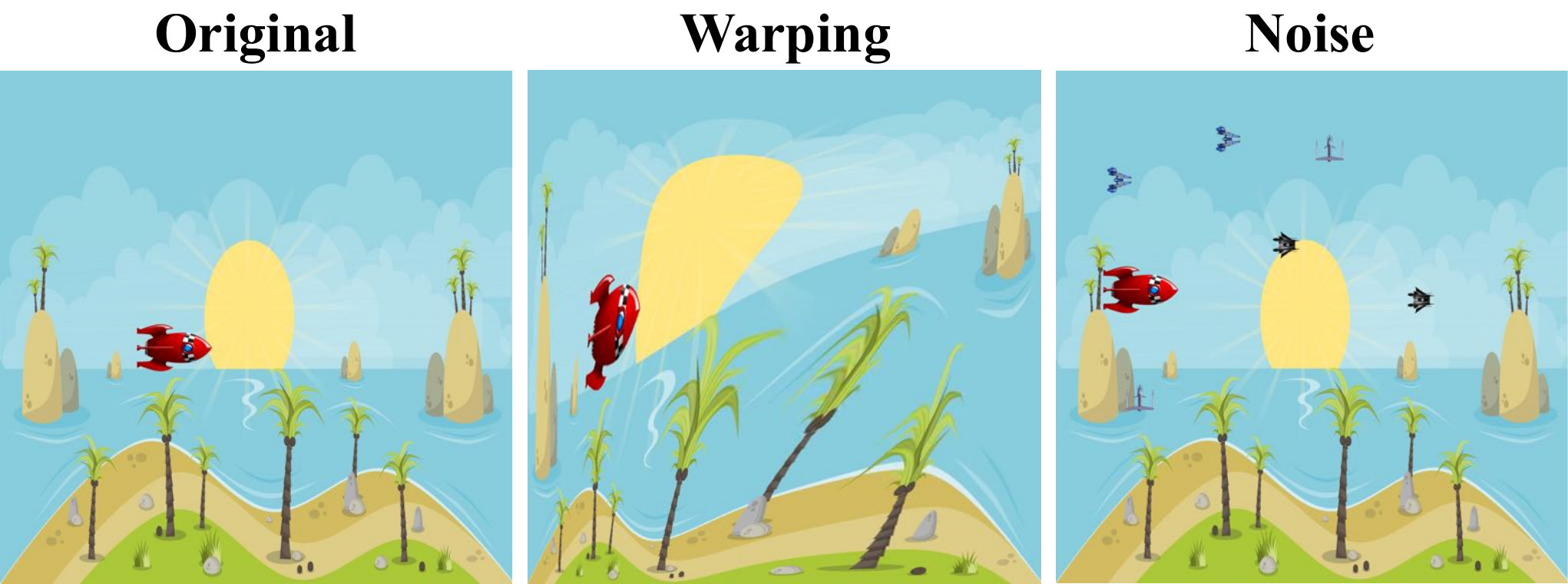}}
\caption{
Our method can discover simple laws of motion even if the true images (left) are severely warped (middle) or 
corrupted by superimposed noise in the form of smaller distractor rockets (right).
\label{warpingFig}
}
\end{figure}

Not only {\it can} our architecture discover unnecessarily complicated solutions, but it by default {\it will}. 
We jocularly termed this the {\it Alexander principle} in honor of a child of one of the authors whose sense of humor dictated that he comply with requests in the most complicated way consistent with the instructions.
We will face multiple challenges of this type throughout this paper, where our neural networks appeared humorously spiteful simply because they statistically find the most {\it generic} solution in a vast class of equally accurate ones.

\def\Lrec{\Ell^{\rm recon}}
\def\Lpred{\Ell^{\rm pred}}
\def\Lnl{\Ell^{\rm nl}}
\def\Lacc{\Ell^{\rm acc}}
\def\Lcurv{\Ell^{\rm curv}}
  
 \def\xrms{x_{\rm rms}}

To tackle this problem, we wish to add a regularization term to the loss function that somehow rewards simplicity and penalizes complexity, ideally in a way that involves as few assumptions as possible about the type of dynamics occurring in the video. 
Defining the 2n-dimensional vector
\beq{wDefEq}
\w_i\equiv \left(\z_{i-2}\atop \z_{i-1}\right)\in\mathbb{R}^{2n},
\eeq
we can view the evolution function $U(\w)$ as a mapping from 
$\mathbb{R}^{2n}$ to $\mathbb{R}^{n}$ that we wish to be as simple as possible.
A natural physics-inspired complexity measure for $U$ is its {\it curvature}
\beq{LcurvDefEq}
\Lcurv\equiv R^{\alpha}_{\mu\nu\beta} R_{\alpha}^{\mu\nu\beta},
\eeq
defined as the squared Riemann tensor that is ubiquitous in differential geometry and general relativity, defined as
\beqa{RiemannDefEq}
R^{\alpha}_{\mu\nu\beta}&\equiv&\Gamma^{\alpha}_{\nu\beta,\mu} -
\Gamma^{\alpha}_{\mu\beta,\nu}
 + \Gamma^{\gamma}_{\mu\beta}\Gamma^{\alpha}_{\nu\gamma} -
\Gamma^{\gamma}_{\nu\beta}\Gamma^{\alpha}_{\mu\gamma},\\
\Gamma^{\alpha}_{\mu\nu}&\equiv& {1\over 2} g^{\alpha\sigma}
\left(g_{\sigma\mu,\nu} + g_{\sigma\nu,\mu} - g_{\mu\nu,\sigma}\right),\\
{\bf g}&\equiv&\J\J^t,
\eeqa
where $\J$ is the Jacobian of $U$, the matrix $\g$ is the induced metric on the latent space $\mathbb{R}^{n}$, 
indices are raised by multiplying by $\g^{-1}$, commas denote derivatives as in standard tensor notation, and the Einstein summation convention is used. Natural alternatives are the squared Ricci curvature $R^{\mu\nu} R_{\mu\nu}$ or the scalar curvature $R \equiv g^{\mu\nu}R_{\mu\nu}$, 
where $R_{\mu\nu}\equiv R^{\alpha}_{\mu\alpha\nu}$.

Unfortunately, these curvature measures are numerically cumbersome, since they require taking 3rd derivatives of the neural-network-defined function $U$ and the Riemann tensor has $n^4$ components. Fortunately, we find that a simpler measure of complexity performs quite well in practice, as reflected by the following loss function:
\beq{LossDefEq}
\mathcal{L}\equiv \Lrec+\alpha\Lpred+\beta\Lnl+\gamma\Lacc.
\eeq
These four terms are averages over all time steps $i$ of the following dimensionless functions:
\beqa{LDefEqs}
\Lrec_i&\equiv& {|\x_i-D(E(\x_i))|\over|\x_i|},\\
\Lpred_i&\equiv& {|\z_i-U(\z_{i-2},\z_{i-1}))|\over |\z_{i-1}-\z_{i-2}|},\label{LpredEq}\\
\Lnl_i&\equiv&{1\over 4n^3}  {|\z_{i-1}-\z_{i-2}|\>\>||\nabla\J(\w_i))||_1},\label{LnlEq}\\
\Lacc_i&\equiv&{1\over n}  ||U(\w_i)-\M\w_i||_1,
\eeqa
$\alpha$, $\beta$, $\gamma$ are tunable hyperparameters, and $n$ is the dimensionality of the latent space.
Here $\Lrec$ is the {\it reconstruction error}, 
$\Lpred$ is the {\it prediction error}, and both 
$\Lnl$ and $\Lacc$ are measures of the complexity of $U$.
$\Lnl$ is a measure of the {\it nonlinearity} of the mapping $U$, since its Jacobian $\J$ will be constant if the mapping is linear. Note that
$\Lnl=0$ implies that $\Lcurv=0$, since if $\J$ is constant, then $\Gamma^{\alpha}_{\mu\nu}=0$ and the curvature vanishes.
Physically, $\Lnl=0$ implies that the dynamics is described by 
coupled linear difference equations, which can be modeled by 
coupled linear differential equations and encompass behavior such as
helical motion in magnetic fields, sinusoidal motion in harmonic oscillator potentials and parabolic motion under gravity. 
$\Lacc$ is a measure of the predicted {\it acceleration}, 
since there is no acceleration if the mapping is $U(\w)=\M\w$,
where
\beq{MdefEq}
\M\equiv
\left(
\begin{tabular}{cc}
-\I	&2\I
\end{tabular}
\right),
\eeq
and $\I$ is the $n\times n$ identity matrix. For example, $x_i=2x_{i-1}-x_{i-2}$ gives uniform 1D motion (with \textit{i} indicating the time step at which the \textit{x} coordinate is recorded).
An alternative implementation not requiring Jacobian gradient evaluation would be
$\Lnl_i\equiv{1\over 2n^2} {||\J(\w_{i+1})-\J(\w_{i})||_2^2}$, and 
an alternative acceleration penalty would be
$\Lacc_i\equiv{1\over n} |U(\zero))|^2+{1\over 2 n^2} ||\M-\J(\w_i))||_2^2$.

\section{Results}
\label{ResultsSec}

\subsection{Latent space learning}

\begin{figure*}[phbt]
\centerline{\includegraphics[width=\linewidth]{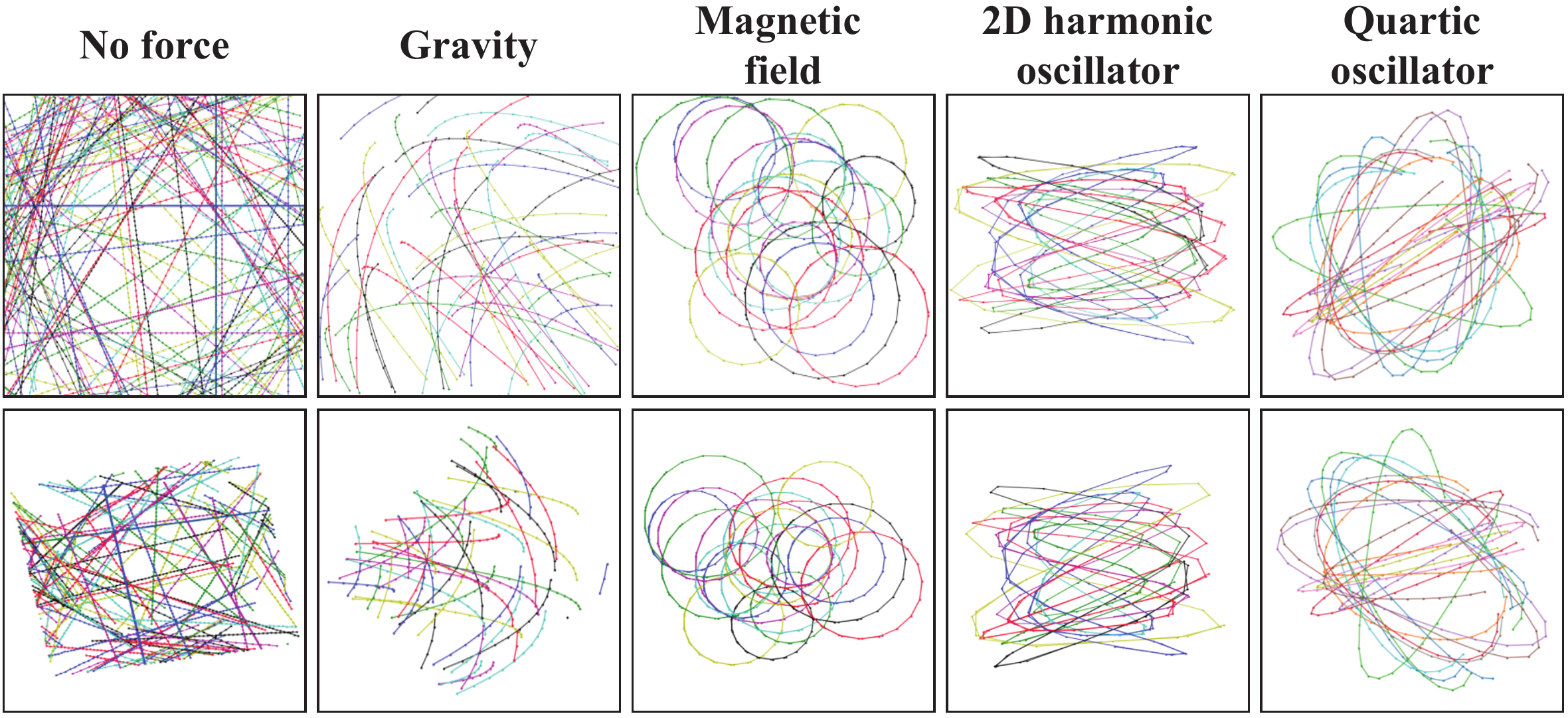}}
\caption{Example of original (top) and discovered (bottom) trajectories in the latent spaces. In the top panel, each point represents the $x$- and $y$-coordinates of the rocket in each frame. In the bottom panel, each point corresponds to the two main principal components discovered in the 5D latent space. In both cases, points of the same color and connected by a line belong to the same trajectory.
\label{latentFig}
}
\end{figure*}

We first tested our algorithm for four physical systems obeying linear differential equations, corresponding to motion with no forces, 
in a gravitational field, in a magnetic field and in a 2D harmonic oscillator potential (see \fig{latentFig}).
For each type of motion, we generated between 100 and 150 trajectories, with around 30 video frames each, corresponding to equally spaced, consecutive time steps. For each trajectory video, the shape of the rocket and the background were kept fixed, but the position of the rocket was changed according to the corresponding physical law of motion, starting with a random initial velocity and a random initial position within the image boundaries. Our training set thus contains a total of 3000-5000 images for each type of motion;
sample trajectories are shown in \fig{latentFig} (top), where each dot represents the $x$- and $y$-coordinate of the rocket in a given frame and points of the same color connected by a line belong to the same trajectory.
After simulating the trajectories and generating a $1000 \times 1000$ pixel image of each video frame
(\fig{QuarticFig} for an example), we downsampled the image resolution to $64 \times 64$ pixels
before passing them to our neural network. 

The encoder network consists of five convolutional ReLU layers with kernel-size 4 and padding 1, four with stride 2 followed by one with stride 1. At the end there is a fully connected linear layer that reduces the output of the encoder to a vector of size equal to the dimension of the latent space.
The number of channels goes from 3 for the input image to 32, 32, 64, 64 and 256 for the convolutional layers.
The decoder network is a mirror image of the encoder in terms of layer dimensions, with the convolution layers replaced by  deconvolution layers.  
The evolution operator has three fully connected 32-neuron hidden layers with softplus activation function and a linear $n$-neuron output layer.
We implemented these networks using PyTorch using a batch size of 256 and the Adam optimizer.
For these four linear types of motion, we set $\gamma=0$ and $\alpha=\beta=10^{-3}$ and trained for 4,000 epochs with a learning rate of $10^{-3}$,
multiplying $\alpha$ and $\beta$ by 10 after every 1000 epochs. We then trained for $3,000$ additional epochs while dividing the learning rate by 10 every 1,000 epochs. 

Although our algorithm successfully learned useful 2D latent spaces  (\fig{latentFig}, bottom) and
predicted images with 2\% {\rms} relative error that were visually nearly indistinguishable from the truth, 
this required overcoming two separate obstacles.
We initially lacked the factor $|\z_{i-1}-\z_{i-2}|$ in \eq{LpredEq}, so by the Alexander principle, 
the neural network learned to drive the prediction loss $\Lpred$ toward zero by collapsing the
latent space to minuscule size. The  $|\z_{i-1}-\z_{i-2}|$-factor solves this problem by making the prediction loss invariant under latent space rescaling.

\begin{figure*}[phbt]
\centerline{\includegraphics[width=\linewidth]{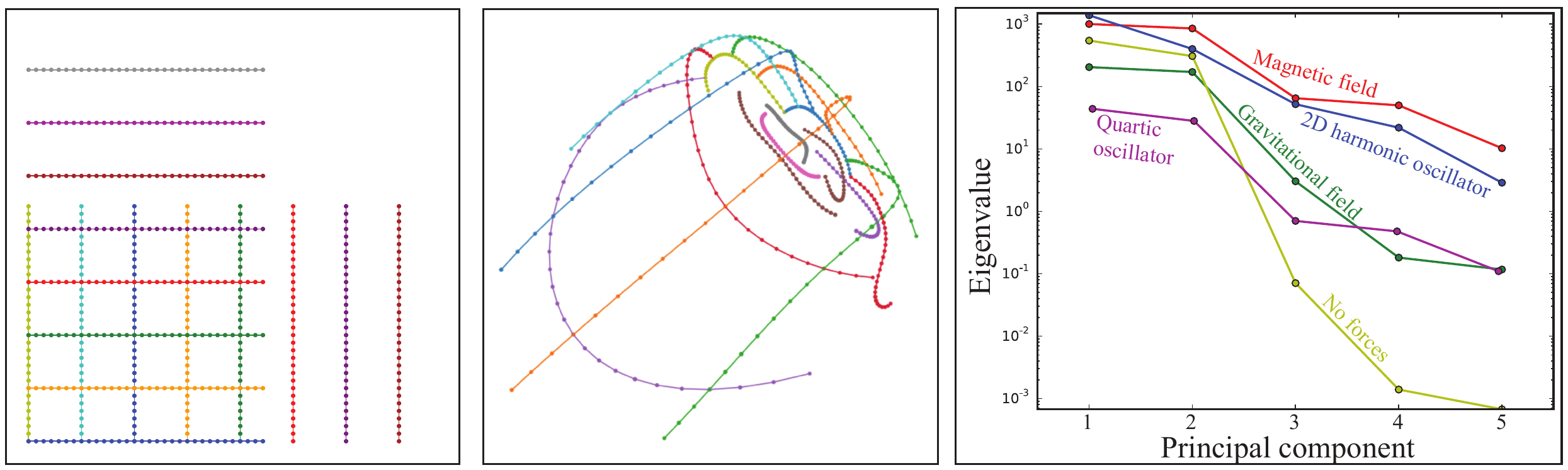}}
\caption{The topological problems (middle) that prevented directly learning a 2-dimensional latent space (left) can be understood via knot theory and eliminated by instead discovering the two main principal components (right) in a learned 5-dimensional latent space.
The left panel shows 16 force-free rocket trajectories, with points denoting the rocket center in each frame and points of the same color corresponds to the same trajectory. The middle panel shows the corresponding points $\z_i$ produced by our encoder network trained with a 2D latent space. The right panel shows the eigenvalues obtained from a PCA on a five-dimensional learned latent space, revealing that the latent space is rather 2-dimensional because two principal components account for most of the variance.
\label{latentProblemFig}
}
\end{figure*}

\subsection{Knot theory to the rescue}

The second obstacle is topological.
If you drop a crumpled-up towel (a 2D surface in 3D space) on the floor, it will not land perfectly flat, but with various folds.
Analogously, the space of all possible rocket images forms a highly curved surface in the $N$-dimensional space of images, so when a randomly initialized neural network first learns to map it into a 2D latent space, there will be numerous folds. For example,  the left panel of \fig{latentProblemFig} shows 16 trajectories (each shown in a different color) corresponding to the rocket moving uniformly in straight lines. The middle panel shows these same trajectories (with the same colors as in the left panel) in the latent space first discovered by our neural network when we allowed only two latent space dimensions.
Some pairs of trajectories which are supposed to be straight parallel lines (left panel) are seen to cross in a cat-like pattern in the latent space (middle panel) even though they should not cross. During training, the network tries to reduce prediction and complexity loss by gradually distorting this learned
latent space to give trajectories the simplest possible shapes (straight lines in this case), but gets stuck and fails to unfold the latent space. This is because the reconstruction loss $\Lrec$ effectively causes distinct images to repel each other in the latent space: if two quite different rocket images get mapped to essentially the same latent-space point, then the decoder will epically fail for at least one. 
Unfolding would require temporarily moving one trajectory across another, thus greatly increasing the loss.
This is analogous to topological defects in physics that cannot be removed because of an insurmountable energy barrier. 

Fortunately, knot theory comes to the rescue: a famous theorem states that there are no $d$-dimensional knots in an $n$-dimensional space if $n>{3\over 2}(1+d)$ 
\cite{zeeman1960unknotting}.
For example, you cannot tie your shoelaces ($d=1$) if you live in $n=4$ dimensions. 
Our topological pregression problem corresponds to the inability of the neural network to untie a $d$-dimensional knot in $n$ dimensions, where $d$ is the dimensionality of the image submanifold of $\mathbb{R}^{N}$ ($d=2$ for our examples). We therefore implemented the following solution, which worked well for all our examples:
First run the pregression algorithm with a latent space of dimension $n'>{3\over 2}(1+d)$ (we found $n' = 5$ to be enough for us) and then
extract an $n$-dimensional latent space using principal component analysis. 
This corresponds to incentivizing the aforementioned towel to flatten out while still in the air and then rotating it to be parallel to the floor before landing.
\fig{latentProblemFig} (right) shows that upon applying PCA to the points in the 5-dimensional latent space, two principal components dominate the rest (accounting for more than $90\%$ of the variance), revealing that 
all rocket images get mapped roughly into a 2D plane (\fig{latentFig}) in a 5D latent space.

\subsection{Nonlinear dynamics and the accuracy-simplicity tradeoff}
\label{non_lin_dyn}

Increasing the two parameters $\beta$ and $\gamma$ in \eq{LossDefEq} penalizes complexity ($\Lnl$ and $\Lacc$) more relative to inaccuracy
($\Lrec$ and $\Lpred$). For our quartic oscillator example (\fig{QuarticFig}), achieving $\Lnl=0$ is impossible and undesirable, since the correct dynamics is nonlinear with $\nabla\J\ne 0$, so we wish to find the optimal tradeoff between simplicity and accuracy.
We did this by training as above for 7,000 epochs but setting $\beta=0$, then keeping $\gamma=\beta$ and further training 14 networks in parallel for a geometric series of $\beta$-values from $0.01$ to $200$. These 14 networks were trained for 3,000 epochs with learning rate starting at $10^{-3}$ and dropping tenfold every 1,000 epochs.

Since, as mentioned above, there is a broad class of equally accurate solutions related by a latent space reparametrization $\z\mapsto f(\z)$, we expect that increasing $\beta$ from zero to small values should discover the simplest solution in this class without decreasing prediction or reconstruction accuracy. This is the solution we want, in the spirit of Einstein's famous dictum {\it ``everything should be made as simple as possible, but not simpler".}
Further increasing $\beta$ should simplify the solution even more, but now at the cost of leaving this equivalence class, reducing accuracy.
Our numerical experiment confirmed this expectation: we could increase regularization to $\beta=50$ (the choice shown in \fig{QuarticFig}) without significant accuracy loss, after which the inacuraccy started rising abruptly. It should be noted that a similar Pareto approach could be used for the other four linear types of motions, but in those cases, the Pareto frontier would be trivial, given that the right solution (minimum loss) corresponds to having no non-linearity (minimum complexity).

\subsection{Image warping and noise}

As mentioned in \Sec{LossSec}, the fact that our algorithm rewards simplicity in the evolution operator $U$ rather than the encoder/decoder pair should enable it to discover the simplest possible latent space even if the space of image $(x,y)$-coordinates is severely distorted.
To test this, we replaced each image with color $c[x,y]$ (defined over the unit square) by a warped image $c'[x,y]$ defined by
\beqa{WarpEq}
c'[x,y]&\equiv&c[g(x) + x (1 - x) y,g(y) + y (1 - y) x],\nonumber\\
g(u)&\equiv&u (11 - 18 u + 12 u^2)/5
\eeqa
as illustrated in \fig{warpingFig} (middle panel), and analyzed the 3,000 warped video frames of the rocket moving in a magnetic field.
As expected, the pregression algorithm recovered a non-warped latent space just as in \fig{latentFig}, so this extra complexity was entirely absorbed
by the decoder/encoder, which successfully learned the warping function $c\mapsto c'$ of \eq{WarpEq} and its inverse.

We also tested the robustness of our pregression algorithm to noise in the form of smaller rockets added randomly to each video frame.
We used 3 different types of distractor rockets as noise, and added between zero and 10 to each image
as illustrated in \fig{warpingFig} (right panel).
The result was that the pregression algorithm learned to reconstruct the latent space just as before, focusing only on the large rocket, and reconstructing images with the distractor rockets removed.

\subsection{Automatically discovering equations and inertial frames}

\begin{figure}[phbt]
\centerline{\includegraphics[width=\linewidth]{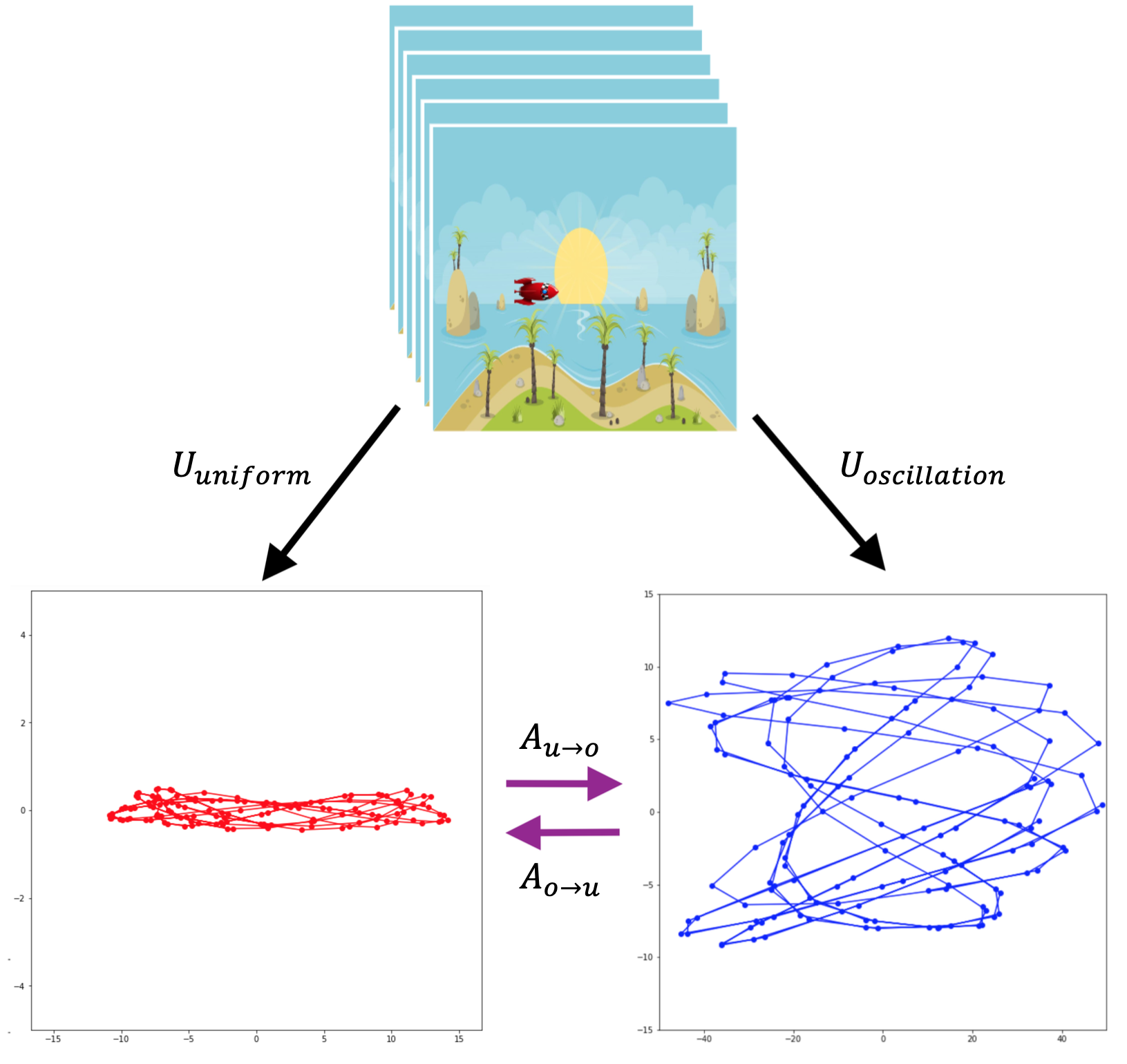}}
\vskip-5mm
\caption{Video trajectories in the 2D harmonic oscillator potential (top) look different when mapped into the latent space using the encoder trained on that same data (right) than when mapped using the encoder trained on uniform force-free motion (left). 
However, the two latent spaces are equivalent up to an affine transformation. 
\label{latSpaceDiff}
}
\end{figure}

\begin{figure*}[phbt]
\centerline{\includegraphics[width=\linewidth]{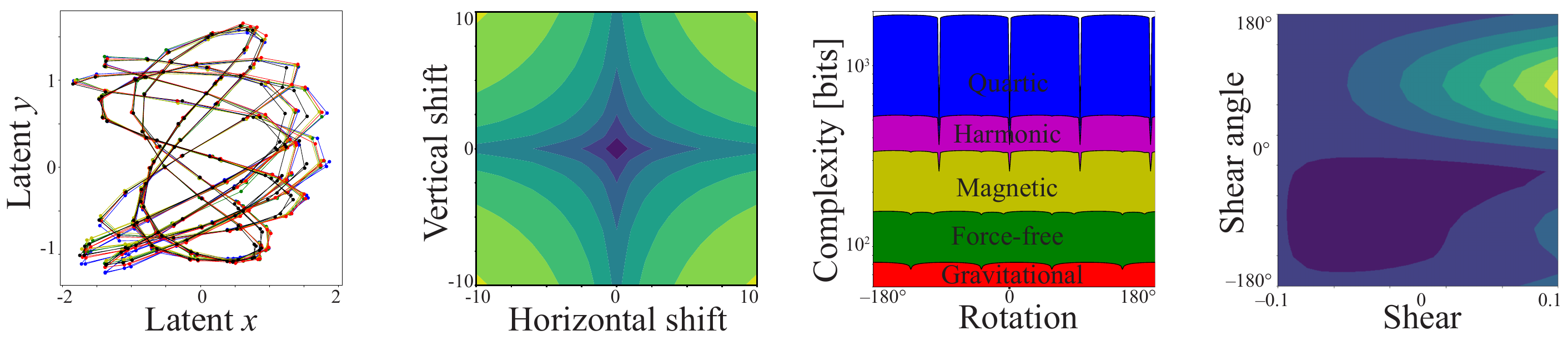}}
\caption{(Left) The five latent spaces can be unified by using affine transformations. The complexity of the equations in this unified space is minimized when having no further shift (2nd panel) or rotation (3rd panel), but a $\sim 3\%$ shear (right).  
\label{affineFig}
}
\end{figure*}

\def\logplus{\log_+}
\def\loss{\ell}
Let us now turn to the task of discovering physical laws that are both accurate and simple. Although the five rocket-motion examples took place in the same image space, the Alexander Principle implies that the five latent spaces (bottom panels in \fig{latentFig}) will generally all be different, since we trained a separate neural network for each case. 
Specifically, we expect the latent spaces to each differ by some affine transformation $\rvec\mapsto\A\rvec+\a$ for some constant vector $\a$ and $2\times 2$ matrix $\A$, since affine transformations do not affect the amount of nonlinearity or acceleration required and thus
 leave our complexity loss functions $\Lacc$ and $\Lcurv$ unchanged.

\Fig{latSpaceDiff} shows the same sequence of 
images of the rocket moving in the 2D harmonic oscillator potential mapped into two latent spaces, learned by
training our pregression algorithm on either that same 2D oscillator data (right panel) or on the force-free data (left panel).
As expected, they trajectories are seen to differ by an affine transformation. 
Indeed, the left panel of  \fig{affineFig} shows that the latent spaces discovered by all our 5 experiments are
interrelated by affine transformations. 
Here we have mapped all latent space coordinates $\rvec_i$, $i=1,...,5$, into a single unified latent space $\rvec=\A_i\rvec_i+\a_i$ by introducing five $2\times 2$ matrices $\A_i$ and 2D translation vectors $\a_i$ to match up corresponding rocket positions (each color is associated with an encoder trained on a specific type of motion: force free (red), 2D oscillator (blue), magnetic (yellow),
gravitational field (green) and quartic oscillatior (black) and green.
Specifically, without loss of generality we take one of the latent spaces (derived from the harmonic oscillator) to the be unified one, so $\rvec_1=\rvec, \A_1=\I$, $\a_1=0$, and solve for the other $\A_i$ and $\a_i$ by 
minimizing the total mismatch
\beq{UnificationEq}
M\equiv \sum_{i=2}^5 \expec{\loss(|\A_i\rvec_i +\a_i-\rvec_1|) + \loss(|\A_i^{-1}(\rvec_1-\a_i) -\rvec_i| )},
\eeq
where the average is over all our rocket images mapped through the the five encoders. 
If the loss function penalizing mismatch distance were $\loss(r)=r^2$, \eq{UnificationEq} would simply be a $\chi^2$-minimization determining $\A_i$ and $\a_i$  via linear regression, except that we have also penalized inaccuracy in the inverse mapping (second term) to avoid biasing $\A_i$ low.
To increase robustness toward outliers, we instead followed the prescription of \cite{wu2019toward} by choosing $\loss(r)\equiv{1\over 2}\log_2\left[1+(r/\epsilon)^2\right]$ and minimizing $M$ with gradient descent, using an annealing schedule $\epsilon=10^1$, $10^0$, ..., $10^{-10}$.


Next, we estimated the velocity $\dot\rvec$ and acceleration $\ddot\rvec$ at each data point by 
cubic spline fitting to each trajectory $\rvec(t)$ in the unified latent space, and discovered candidate differential equations of the form $\ddot\rvec=f(\dot\rvec,\rvec)$ using the publicly available AI Feynman symbolic regression package \cite{udrescu2020ai,udrescu2020ai2}.
To eliminate dependence on the cubic spline approximation, we then
recomputed the accuracy of each candidate formula $f$ by using it to predict each data point from its two predecessors using the boundary-value ODE solver {\it scipy.integrate.solve\_bvp} \cite{2020SciPy-NMeth}, selecting the most accurate formula for each of our five examples.

Applying an affine transformation $\rvec\mapsto\A\rvec+\a$ to both the data and these equations of course leaves the prediction accuracy unchanged, so we now exploit this to further reduce the total information-theoretic complexity of our equations, defined as in \cite{wu2019toward}.
\Fig{affineFig} (2nd panel) shows a contour plot of the equations complexity as a function of an overall shift (darker means smaller). We observe a clear optimum for the shift vector $\a$, corresponding to eliminating additive 
constants in the harmonic and quartic oscillator equations. For example, $\ddot x=-x$ is simpler than 
$\ddot x=2.236-x$. The 3rd panel of \Fig{affineFig} shows the total equation complexity (as a stacked histogram) as a function of an overall rotation of the coordinate axis. We see several minima: the gravitational example likes $45^\circ$ rotation because this makes the new horizontal acceleration vanish, but the other examples outvote it in favor of a $0^\circ$ rotation to avoid $xy$ cross-terms.  

Only three degrees of freedom now remain in our matrix $\A$: shear (expanding along some axis and shrinking by the inverse factor along the perpedicular axis) and an overall scaling.
We apply the {\it vectorSnap} algorithm of \cite{udrescu2020ai2} to discover rational ratios between parameters and then select the shear that maximizes total accuracy. As it can be seen in the contour plot in the right panel of \Fig{affineFig} (darker color means smaller values), $\sim 3\%$ shear is optimal. Finally, we apply the scaling that minimizes total complexity, resulting in these discovered laws of motion for the force-free, gravitational, magnetic, 2D oscillator and quartic oscillator examples, respectively:
\begin{equation}
\begin{split}
&\left({\ddot x\atop\ddot y}\right)=\left({0\atop 0}\right)\\
&\left({\ddot x\atop\ddot y}\right)=-\left({1\atop 1}\right) \\
&\left({\ddot x\atop\ddot y}\right)={1\over 3}\left({\dot y\atop-\dot x}\right) \\
&\left({\ddot x\atop\ddot y}\right)=-{1\over 9}\left({4x\atop y}\right) \\
&\left({\ddot x\atop\ddot y}\right) = -7.3\times 10^{-6}(x^2+y^2)\left({x\atop y}\right)
\end{split}
\end{equation}
These are in fact exactly the laws of motion used to generate our training set images (up to some noise in the quartic term prefactor), but
reexpressed in a five times smaller latent space than the one we used, which further simplifies our formulas (for example, the discovered gravitational acceleration is 1 instead of 5).

\section{Summary}
\label{ConclusionsSec}

We have presented a method for unsupervised learning of equations of motion for objects in raw and optionally distorted unlabeled video.
This automatic un-distortion may be helpful for modeling real-world video afflicted by
stereoscopic projection, lens artifacts, varying lighting conditions, {\etc}, and also for learning degrees of freedom such as 3D coordinates and rotation angles. Our method is in no way limited to video, and can be applied to any 
time-evolving dataset, say $N$ numbers measured by a set of sensors.
Although we focused on dynamics, it could also be interesting to generalize our approach to other situations, by attempting to infer 
other properties of the system rather than its future state.  Another interesting avenue for future work is to explore whether 
the above-mentioned topological intuition provided by knot theory can help improve autoencoders more generally. 

The reparametrization invariance of general relativity teaches us that there is an infinite class of coordinate systems
that provide are equally valid physical descriptions, and we found a similar reparametrization invariance of our auto-discovered latent space. 
We broke this degeneracy by quantifying and minimizing the geometric and symbolic complexity of the dynamics. Although different systems were simplest in different coordinate systems, we found that minimizing total complexity for all of them recovered a standard isotropic inertial frame.
An interesting topic for future work would be to explore whether our brains' representations of physical systems are similarly optimized to make prediction as simple as possible.

{\bf Acknowledgements:}
The authors with to thank Zhiyu Dong, Jiahai Feng, Bhairav Mehta, Andrew Tan and Tailin Wu for helpful comments,  and the Center for Brains, Minds, and Machines (CBMM) for hospitality.
This work was supported by Institute for AI and Fundamental Interactions, The Casey and Family Foundation, the Ethics and Governance of AI Fund, the Foundational Questions Institute, the Rothberg Family Fund for Cognitive Science and the Templeton World Charity Foundation, Inc.

\bibliographystyle{unsrt}
\bibliography{pregression}

\end{document}